\documentclass[10pt,twocolumn,letterpaper]{article}

\usepackage{wacv}
\usepackage{times}
\usepackage{epsfig}
\usepackage{graphicx}
\usepackage{amsmath}
\usepackage{amssymb}

\usepackage{multirow}

\usepackage[pagebackref=true,breaklinks=true,letterpaper=true,colorlinks,bookmarks=false]{hyperref}

\wacvfinalcopy 

\def\wacvPaperID{****} 

\ifwacvfinal\fi
\begin{document}

\title{Silhouette Guided Point Cloud Reconstruction beyond Occlusion}

\author{Chuhang Zou ~~~~~~~~~~~~~~~~Derek Hoiem\\
University of Illinois at Urbana-Champaign\\
{\tt\small \{czou4, dhoiem\}@illinois.edu}
}

\maketitle
\ifwacvfinal\thispagestyle{empty}\fi

\begin{abstract}
One major challenge in 3D reconstruction is to infer the complete shape geometry from partial foreground occlusions. In this paper, we propose a method to reconstruct the complete 3D shape of an object from a single RGB image, with robustness to occlusion. Given the image and a silhouette of the visible region, our approach completes the silhouette of the occluded region and then generates a point cloud. We show improvements for reconstruction of non-occluded and partially occluded objects by providing the predicted complete silhouette as guidance. We also improve state-of-the-art for 3D shape prediction with a 2D reprojection loss from multiple synthetic views and a surface-based smoothing and refinement step. Experiments demonstrate the efficacy of our approach both quantitatively and qualitatively on synthetic and real scene datasets.
   
\end{abstract}

\section{Introduction}
3D reconstruction from 2D images has many applications in robotics and augmented reality. One major challenge is to infer the complete shape of a partially occluded object. Occlusion frequently occurs in natural scenes: \eg we often see an image of a sofa occluded by a table in front and a dining table partially occluded by a vase on top. Even multi-view approaches~\cite{ulusoy2017semantic, guan2010multi, kang2001handling} may fail to recover complete shape, since occlusions may block most views of the object. Single-view learning-based methods~\cite{guo2015labeling, ehsani2018segan, zhu2017semantic} have approached seeing beyond occlusion as a 2D semantic segmentation completion task, but complete 3D shape recovery adds the challenges of predicting 3D shape from a 2D image and being robust to the unknown existence and extent of an occluding region.  

\begin{figure}[t]
\begin{center}
\includegraphics[width=1.0\linewidth]{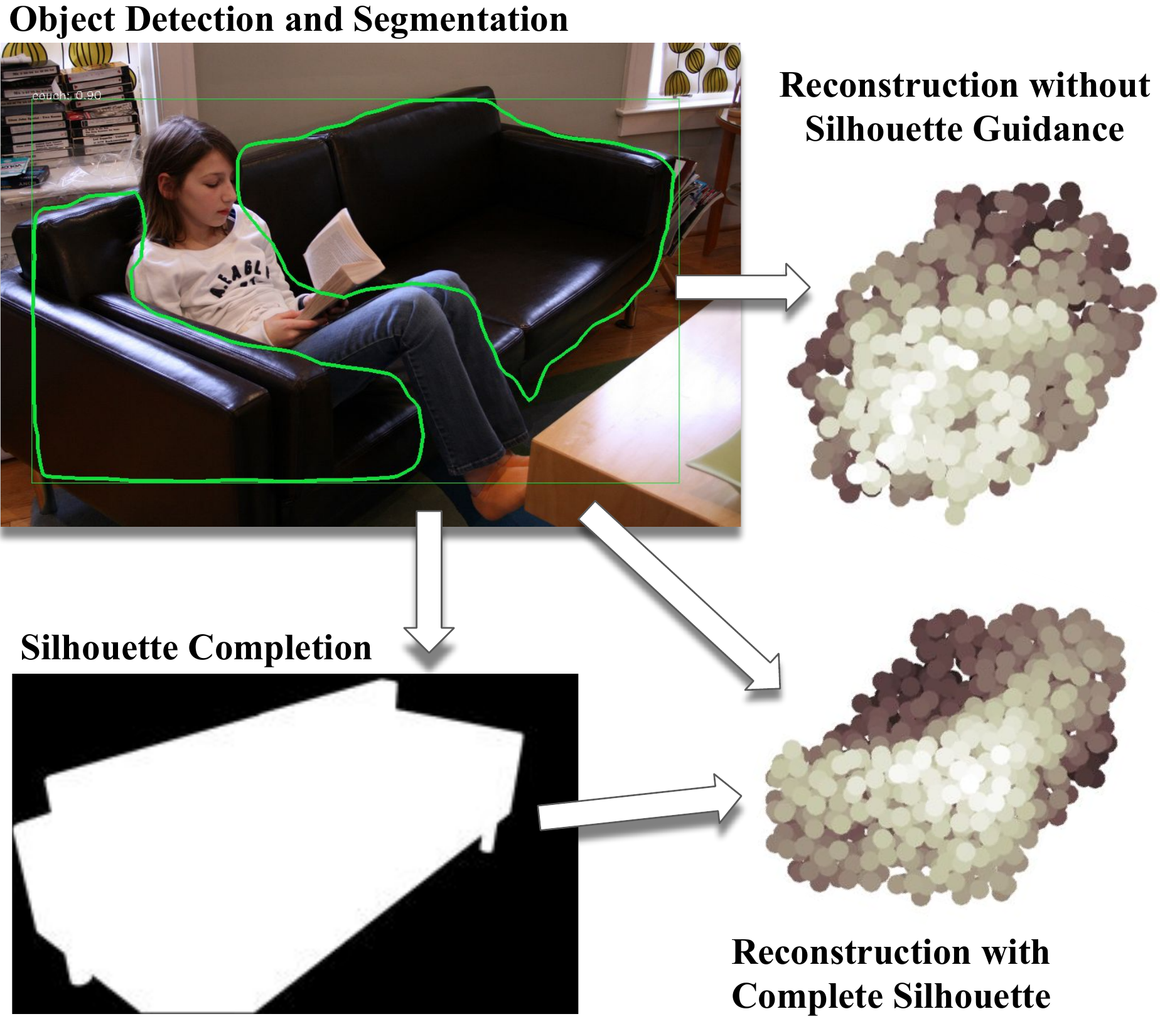}
\end{center}
\vspace{-3mm}
   \caption{\textbf{Illustration.} The person sitting on the sofa blocks much of the sofa from view, causing errors in existing shape prediction methods. We propose improvements to shape prediction, including the prediction and completion of the object silhouette as an intermediate step, and demonstrate more accurate reconstruction of both occluded and non-occluded objects. Best viewed in color.}
   \vspace{-2mm}
\label{fig:illustration}
\end{figure}

In this paper, our goal is to reconstruct a complete 3D shape from a single RGB image, in a way that is robust to occlusions.  We follow a data-driven approach, using convolution neural networks~(CNNs) to encode shape-relevant features and decode them into an object point cloud. To simplify the shape prediction, we split the task into: (1) determining the visible region of the object; (2) predicting a completed silhouette (filling in any occluded regions); and (3) predicting the object 3D point cloud based on the silhouette and RGB image~(Fig.~\ref{fig:illustration}). We reconstruct the object in a viewer-centered manner, inferring both object shape and pose.  We show that, provided with ground truth silhouettes, shape prediction achieves nearly the same performance for occluded objects as non-occluded objects.
We obtain the visible portion of the silhouette using Mask-RCNN~\cite{he2016deep} and then predict the completed silhouette using an auto-encoder.  Using the predicted silhouette as part of shape prediction also yields large improvements for both occluded and non-occluded objects, indicating that providing an explicit foreground/background separation for the object in RGB images is helpful~\footnote{Code and data is available at \url{https://github.com/zouchuhang/Silhouette-Guided-3D}}.  



\begin{figure*}
\begin{center}
\includegraphics[width=1.0\linewidth]{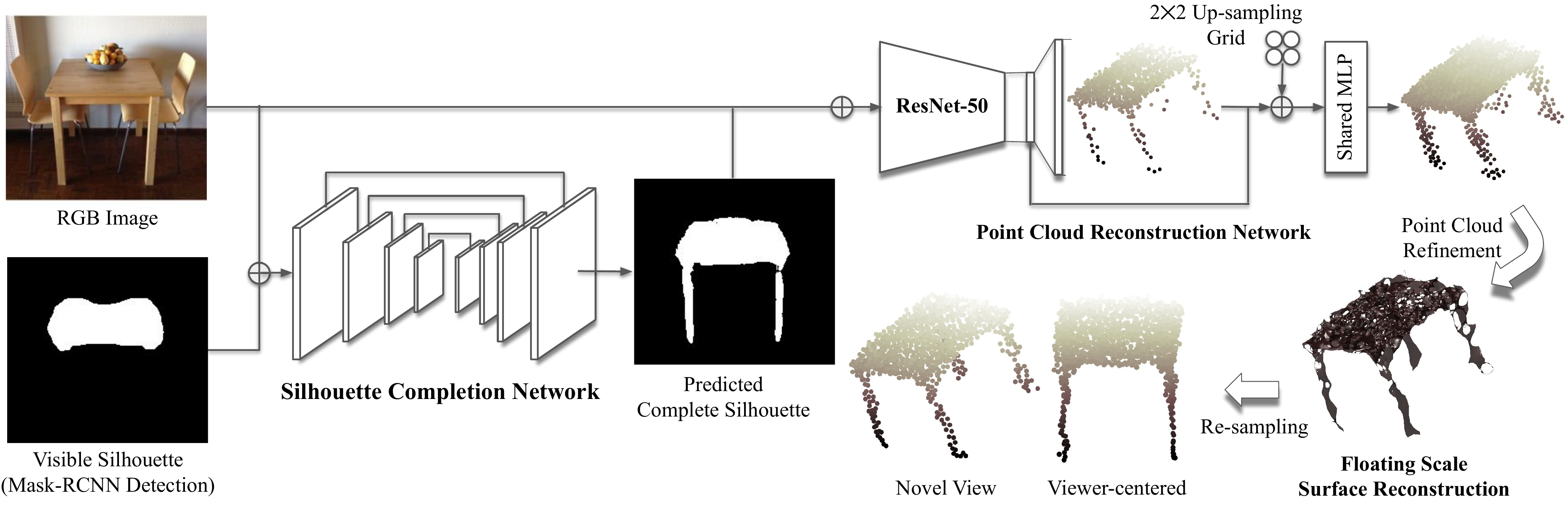}
\end{center}
\vspace{-3mm}
   \caption{\textbf{Approach}. Our approach is composed of three steps. In the first step, the silhouette completion network takes an RGB image and the visible silhouette as input, and predict the complete silhouette of the object beyond foreground occlusions. In the second step, given the RGB image and the predicted complete silhouette, the reconstruction network predicts point clouds in viewer-centered coordinates. Finally, we perform a post refinement step to produce smooth and uniformly distributed point clouds. Best viewed in color.}
   \vspace{-2mm}
\label{fig:overview}
\end{figure*}

Our reconstruction represents a 3D shape as a set of point clouds, which is flexible and easy to transform. Our method follows an encoder-decoder strategy, and we demonstrate performance gains using a 2D reprojection loss from multiple synthetic views and a surface-based post refinement step, achieving state-of-the-art. 
Our silhouette guidance approach is related to shape from silhouette~\cite{karsch2013boundary, barron2012color, koenderink1984does}, but our silhouette guidance is part of learning approach rather than explicit constraint. 

Our contributions:
\begin{itemize}
\vspace{-3mm}
\item We 
improve the state-of-the-art for 3D point clouds reconstruction from a single RGB image. We show performance gains by using a 2D reprojection loss on multiple synthetic views and a surface-based refinement step.\\ 
\vspace{-6mm}
\item 
We demonstrate that completing the visible silhouette leads to better object shape completion. We propose a silhouette completion network that achieves the state-of-the-art. We show improvements for reconstruction of non-occluded and partially occluded objects.
\end{itemize}


\section{Related Work}
\textbf{Single image 3D shape reconstruction} is an active topic of research. Approaches use RGB images~\cite{wang2018pixel2mesh, wu2016single, wang2018adaptive, groueix2018papier, shin2018pixels}, depth images~\cite{yang2018dense, zou20173d, wu20153d, rock2015completing} or both~\cite{gupta2015aligning, fowler2017towards, gupta2015indoor}. 
Approaches include exemplar based shape retrieval and alignment~\cite{bansal2016marr, aubry2014seeing, gupta2015aligning, izadinia2017im2cad}, deformations from meshes~\cite{kar2015category, kong2017using, wang2018pixel2mesh}, or a direct prediction via convolution neural networks~\cite{wu2016single,yan2016perspective,wu2017marrnet}. Qi~\etal~\cite{qi2017pointnet} propose a novel deep net architecture suitable for consuming unordered point sets in 3D; Fan~\etal~\cite{fan2017point} propose to generate point clouds from a single RGB image using generative models. More recent approaches improve point set reconstruction performance by learning representative latent features~\cite{mandikal20183dlmnet} or by imposing constraints of geometric appearance in multiple views~\cite{jiang2018gal}. 

Most of these approaches are applied to non-occluded objects with clean backgrounds and no occlusions, which may prevent their application to natural images.
Sun~\etal~\cite{sun2018pix3d} conduct experiments on real images from Pix3D, a large-scale dataset with aligned ground-truth 3D shapes, but do not consider the problem of occlusion. We are concerned with predicting shape of objects in natural scenes, which may be partly occluded. Our approach improves the state-of-the-art for object point set generation, and is extended to reconstruct beyond occlusion with the guidance of completed silhouettes. 
Our silhouettes guidance is closely related to the human depth estimation by Rematas~\etal~\cite{rematas2018soccer}. However, Rematas~\etal use the visible silhouette~(semantic segmentation) rather than a complete silhouette, making it hard to predict overlapped~(occluded) regions. Differently, our approach conditions on predicted silhouette to resolve occlusion ambiguity, and is able to predict complete 3D shape rather than 2.5D depth points.


\textbf{Seeing beyond occlusion.}  Occlusions have long been an obstacle in multi-view reconstruction. Solutions have been proposed to recover portions of surfaces from single views, \eg with synthetic apertures~\cite{vaish2006reconstructing,favaro2003seeing}, or to otherwise improve robustness of matching and completion functions from multiple views~\cite{ulusoy2017semantic, guan2010multi, kang2001handling}. Other work decompose a scene into layered depth maps from RGBD~\cite{liu2016layered} images or video~\cite{zitnick2004high} and then seek to complete the occluded portions of the maps.  But errors in layered segmentation can severely degrade the recovery of the occluded region. Learning-based approaches~\cite{guo2015labeling, ehsani2018segan, zhu2017semantic} have posed recovery from occlusion as a 2D semantic segmentation completion task. Ehsani~\etal~\cite{ehsani2018segan} propose to complete the silhouette and texture of an occluded object. Our silhouette completion network is most similar to Ehsani~\etal, but we ease the task by predicting the complete silhouette rather than the full texture. We demonstrate better performance with our up-sampling based convolution decoder instead of fully connected layers used in Ehsani~\etal. Moreover, We go further to try to predict the complete 3D shape of the occluded object.


\section{Point Clouds from a Single RGB Image}~\label{text:pts}
Direct point set prediction from a single image is challenging due to the unknown camera viewpoint or object pose and large intraclass variations in shape.
 This requires a careful design choice on the network architecture. We aim to have an encoder that can capture object pose and shape from a single image, and a decoder that is flexible in producing unordered, dense point clouds. 

In this section, we introduce our point prediction network architecture, the training scheme including a 2D reprojection loss on multiple synthetic views to improve performance~(Sec.~\ref{text:pt_train}). We then introduce a post-refinement step via surface-based fitting~(Sec.~\ref{text:fssr}) to produce smooth and uniformly distributed point sets.

\subsection{Point Cloud Reconstruction Network}\label{text:pt_train}
Our network architecture is illustrated in Fig.~\ref{fig:overview}. The network predicts 3D point clouds in viewer-centered coordinates. 
The encoder is based on ResNet-50~\cite{he2016deep} to better capture object shape and pose feature. The decoder follows a coarse-to-fine multi-stage generation scheme in order to efficiently predict dense points with limited memory. Our decoder follows the design of PCN~\cite{yuan2018pcn}. The coarse predictor predicts $N=1024$ sparse points. The refinement branch produces 4$N$ finer points, by learning a $2\times2$ up-sampling surface grid centered on each coarse point via local folding operation. 
We experimented with a higher up-sampling rate (\eg 9, 16) as PCN but observed repetitive patterns across all surface patches, missing local shape details.  
Note that our network is able to generate a denser prediction with another up-sampling branch on top, but the current structure best balances accuracy and training/inference speed. Our reconstruction network does not require features from partial points like PCN, and produces an on-par performance with PSG~\cite{fan2017point}, a state-of-the-art method in point set generation~(see experiments in Sec.~\ref{exp:svr}), even without the refinement step we will introduce in Sec.~\ref{text:fssr}.

\textbf{Loss function.} We consider the training loss in 3D space using the bidirectional Chamfer distance. Given predicted point clouds $\hat{p}\in\hat{P}$ and the ground truth $p\in P$, we have:
\begin{align}\label{loss:3d}
    L_{\mathit{rec}}&=d_{\mathit{Chamfer}}(P, \hat{P})\nonumber\\
    &=\frac{1}{|P|}\sum_{p\in P}\min_{\hat{p}\in\hat{P}}\|p-\hat{p}\|_2^2+\frac{1}{|\hat{P}|}\sum_{\hat{p}\in \hat{P}}\min_{p\in P}\|p-\hat{p}\|_2^2
\end{align}
To further boost the performance, we propose a 2D reprojection loss on point sets as follows: 
\begin{align}
    L_{\mathit{proj}} &= d (\mathit{Proj}(P), \mathit{Proj}(\hat{P})) \nonumber\\
    & = \frac{1}{|P|}\sum_{p\in P}\min_{\hat{p}\in\hat{P}}\|K[R~t]p-K[R~t]\hat{p}\|_2^2~\label{loss:2d}
\end{align}
Where $\mathit{Proj}(\cdot)$ is a 2D projection operation from 3D space, with 3D rotation $R$ and translation $t$ in world coordinates and a known camera intrinsic $K$. Since our reconstruction is viewer-centered, we can simply set $R = I, t=0$ assuming projections on the image plane. Our 2D reprojection loss is an unidirectional Chamfer distance; we only penalize the 
average distance from each projected ground truth point to the nearest projection of predicted point cloud. This is because the Chamfer distance on another direction tends to be redundant. When the predicted point is projected inside the ground truth 2D segmentation, the distance to the nearest projected ground truth points tends to zero, resulting in a small gradient and having less effect for learning better 3D point clouds. 
Although we project points instead of surfaces or voxel occupancy, producing non-continuous 2D segments, our 2D reprojection loss is computational efficient and shows promising improvements in experiment. Moreover, fitting a surface for post-refinement~(Sec.~\ref{text:fssr}) to these points is effective. 

We can extend Eq.~\ref{loss:2d} to project 3D points onto multiple orthographic synthetic views: \eg projecting to $x-y$ plane, $y-z$ plane~(image plane) or $x-z$ plane in world coordinates~(detailed illustration in Appx.~\ref{appx:proj}). 
In this case we can simply change the rotation matrix $R$ based on each view. 
Our 2D reprojection loss does not require additional rendered 2D silhouette ground truth of known view points, which makes the training possible on the dataset where the 3D ground truth is available. 

When being projected on the $y-z$ plane~(image plane), the ground truth is a subset of the ground truth silhouette. We thus use 2D points sampled from the ground truth segmentation mask $S$ instead of projecting ground truth 3D points $P$:
\begin{align}
    L_{\mathit{silhouette}} 
     = \frac{1}{|S|}\sum_{s\in S}\min_{\hat{p}\in\hat{P}}\|s-K[R~t]\hat{p}\|_2^2
\end{align}\label{loss:2dd}


The \textbf{overall loss function} is shown below:
\begin{equation}
    L = w_\mathit{rec}L_{\mathit{rec}}+w_\mathit{silhouette}L_{\mathit{silhouette}}+w_\mathit{proj}L_{\mathit{proj}}\label{loss:all}
\end{equation}
which is the weighted summation over the 3D Chamfer loss and the 2D reprojection losses. Here $L_{\mathit{silhouette}}$ is the 2D reprojection loss on the image plane, and $L_{\mathit{proj}}$ is the projection on $x-y,x-z$ planes. 
Note that different from PCN, our network only penalizes on the finest output, which helps ease training and shows no performance degrades. 

\textbf{Implementation details.} Our network gets as input a $224\times 224$ image with pixel values normalized to $[0,1]$. 
The bottleneck feature size is 1024. The coarse decoder consists two fc layers, with feature size of 1024 and 3072 and ReLU in between. We set the surface grid for point up-sampling to be zero-centered with a side length of 0.1. We use the ResNet encoder pre-trained from ImageNet and apply a stage-wise training scheme for faster convergence and easier training: first train to predict coarse point cloud, fix the trained layers, then train the up-sampling header, and finally train the whole network end-to-end. We use ADAM~\cite{kingma2014adam} to update network parameters with a learning rate of $1e^{-4}$ and $\epsilon=1e^{-6}$ and batch size 32. We set $w_\mathit{rec}=1$, $w_\mathit{silhouette}=1e^{-9}$ and $w_\mathit{proj}=1e^{-10}$ in Eq.~\ref{loss:all} based on grid search in the validation set.

\textbf{Data augmentation.} We augment the training samples by gamma correction with $\gamma$ between 0.5-2. We re-scale image intensity with a minimum intensity ranges between 0-127 and a fixed maximum intensity of 255. We add color jittering to each RGB channel independently by multiplying a factor ranges in 0.8-1.2. Each augmentation parameter is uniformly and randomly sampled from the defined range.

\begin{figure}[t]
\begin{center}
\includegraphics[width=1.0\linewidth]{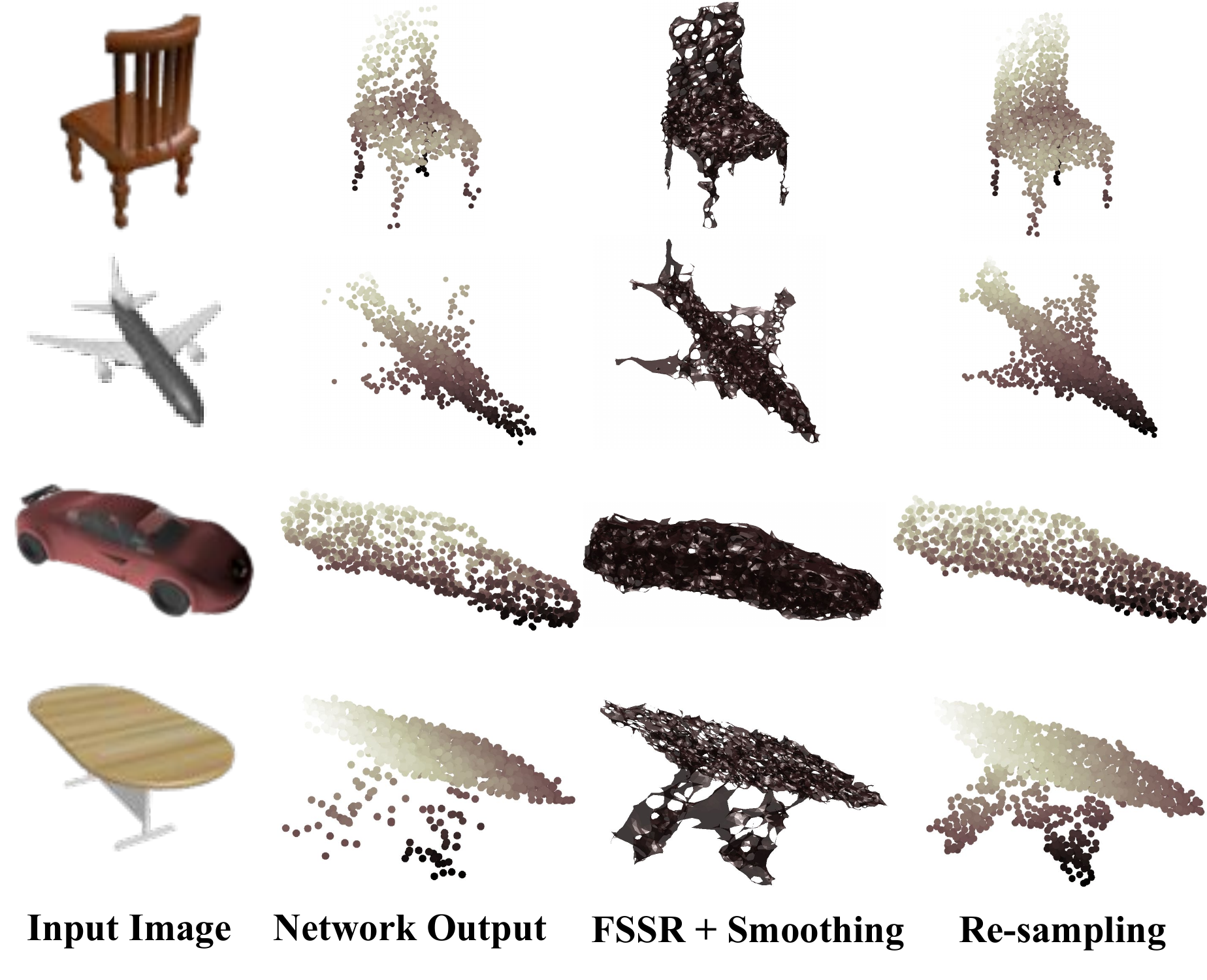}
\end{center}
\vspace{-4mm}
   \caption{Surface-based point clouds refinement. We show from left to right: input RGB image, network prediction, FSSR surface fitting and smoothing and the re-sampled point clouds from the fitted surfaces. Each sample consists the same number of points and we visualize each predicted shape in a novel view for better illustration. Our refinement step is able to produce smooth and uniformly distributed point sets. Best viewed in color.}
   \vspace{-2.5mm}
\label{fig:fssr}
\end{figure}

\subsection{Surface-based Point Clouds Refinement}\label{text:fssr}
One important 3D shape property is the smooth and continuous shape surfaces, especially for thin structures like chair legs and light stands. To impose this property, we perform a post-refinement step~(Fig.~\ref{fig:fssr}), fitting surfaces from dense points, smoothing surfaces and uniformly re-sampling points from the surfaces again. Our surface fitting method is based on Floating Scale Surface Reconstruction (FSSR)~\cite{fuhrmann2014floating}, which is the state-of-the-art for surface-based reconstruction from dense point clouds. 
We set the parameter of point-wise normal by plane fitting on 6 neareast neighbors, and set the per-point scale as the average distance to the two closest points. 
A mesh cleaning step goes after FSSR to remove small and redundant patches. We also experimented with Poisson surface reconstruction~\cite{kazhdan2006poisson}, but FSSR produces a better surface in our case. 

\textbf{Smoothing.} Given the fitted surfaces, we perform smoothing by implicit integration method~\cite{desbrun1999implicit}. We use curvature flow as the smoothing operator for 5 iterations. 
We then uniformly sample point clouds based on Poisson disc sampling to obtain our final output. 


Our surface fitting, smoothing, and re-sampling enables production of evenly distributed points that can model thin structures and smooth surfaces, but the predicted shape may not be closed, preventing volumetric-based evaluation. Also, small and disconnected sections of points that model details may be lost, and points that are incorrectly connected to the mesh surfaces may increase errors in some area.  Overall, though, our proposed post-processing refinement step improves the performance~(
large gain in earth-mover's distance with a small cost to Chamfer distance).


\section{Reconstruction of Occluded Objects}
So far, our proposed point cloud reconstruction approach does not consider foreground occlusions: such as a table in front of a sofa, or a person or pillows on the sofa.  Standard approaches do not handle occlusions well since the model does not know whether or where an object is occluded. 
Starting with an RGB image and an initial silhouette of the visible region, which can be acquired by recent approaches such as Mask-RCNN~\cite{he2016deep}, we propose a 2D silhouette completion approach to generate the complete silhouette of the object.  We show that prediction based on the true completed silhouette greatly improves shape prediction and brings performance on occluded objects close to non-occluded; using predicted silhouettes also improves performance, with completed silhouettes outperforming predicted silhouettes of the visible portion.


\subsection{Silhouette Completion Network}\label{text:slc_network}
We assume a detected object, its RGB image crop $I$ and the segmentation of the visible region $S_v$. 
Our silhouette completion network~(Fig.~\ref{fig:overview}) predicts the complete 2D silhouette $S_f$ of the object based on $S_v$. The network follows an encoder-decoder strategy, gets as input the concatenation of $I$ and $S_v$, and predicts $S_f$ with the same resolution as $S_v$. The encoder is a modified ResNet-50 and the decoder consists 5 up-sampling layers, producing a single channel silhouette. Intuitively, when occlusion occurs, we want the network to complete silhouette, and when no occlusion occurs, we want the network to predict the original segmentation. We add skip connections to obtain this property. We concatenate the feature after the $i-th$ conv layer of the encoder to the input feature layer of the $(6-i)-th$ decoder layer. A full skip connection helps ease training and produces the best results. 

\textbf{Implementation details.} For each input image $I$ and visible silhouette $S_v$, we resize them to $224\times 224$ with preserved aspect ratio and white pixel padded. The value of $I$ is re-scaled to $[0,1]$ and $S_v$ is a binary map with 1 indicating the object. For the encoder, we remove the top fc layer and the average pooling layer of a pre-trained ResNet on ImageNet and obtain a bottleneck feature of $7\times 7$. The decoder applies nearest neighbor up-sampling with a scale factor of 2, followed by a convolution layer~(kernel size $3\times3$, stride 1) and ReLU. The decoder feature sizes are 2048, 1024, 512, 256, 64 and 1 respectively, with a Sigmoid operation on top. We train the network using binary cross entropy loss between the prediction and the ground truth complete silhouette. We use ADAM to update network parameters with a learning rate of $1e^{-4}$ and $\epsilon=1e^{-6}$ and batch size 32. Our final prediction is a binary mask obtained by a threshold of 0.5. To account for the truncation of the full object due to the unknown extent of occluded region, we expand the bounding box around each object by 0.3 on each side.

\textbf{Data augmentation.} For each training sample, we perform random cropping on the input image. We crop with a uniformly and randomly sampled ratio ranges between $[0.2,0.4]$ on each side of the input image. 
Other augmentations include left-right flipping with 50\% probability and random rotation uniformly sampled within $\pm5$ degrees on the image plane. We also perform image gamma correction, intensity changes and color jittering as in Sec.~\ref{text:pt_train}. 

\subsection{Silhouette Guided 3D Reconstruction}\label{text:occ_network}
Given the predicted complete silhouette $S_f$, we modify our point cloud reconstruction network to be robust to occlusion. We concatenate our predicted complete silhouette $S_f$ as an additional input channel to the input RGB image $I$, which can effectively guide reconstruction for both partially occluded and non-occluded objects. We show in experiments the importance of using silhouette compared to the approach with no silhouette guidance at all.  

\textbf{Synthetic occlusion dataset.} Since there is no large-scale 3D dataset of rendered 2D object image with occlusion and the complete silhouette ground truth available, we propose to generate a synthetic occlusion dataset. Instead of off-line rendering samples which is time consuming and has limited variety of occlusion, we propose a ``cut-and-paste'' strategy to create random foreground occlusion. Starting with a set of pre-rendered 2D images without occlusion and with known ground truth silhouettes, for each input image $I$, 
we randomly select another image $I'$ from the same split~(train/val/test) of the dataset as $I$, cutting out the object segment $O'$ from $I'$, pasting and overlaying $O'$ on the input image $I$. To be more specific, we paste on the location uniformly sampled from $[(h_0-h', w_0-w'), (h_1+h', w_1+w')]$, where $[(h_0, w_0), (h_1, w_1)]$ denotes the top left and bottom right position of the bounding box around the object segment $O$ in $I$. $h', w'$ are the height and width of the pasted segment $O'$ which is considered as foreground occlusion. 
To ease training, we exclude input samples with pasted occlusion covering over 50\% of the complete object segment. We perform ``cut-and-paste'' with 50\% probability in training and further add randomly sampled real-scene background with 50\% probability. We use the same data augmentation as in Sec.~\ref{text:pt_train} to train the network. We penalize the network on the ground truth complete 3D point clouds and the 2D reprojection loss assuming full shape 2D reprojection. We use the ground truth visible and complete silhouettes to train the network.  


\section{Experiments}
\subsection{Setup}

We verify the following three aspects of our proposed framework: (1) the performance of our reconstruction network compared with the state-of-the-art, the positive impact of surface refinement and reprojection loss~(Sec.~\ref{exp:svr}); (2) the performance of our silhouette completion network compared with the state-of-the-art~(Sec.~\ref{exp:sc}); (3) the impact of silhouette guidance with robustness to occlusion~(Sec.~\ref{exp:occ}). Please find more results in the appendix.

We use ShapeNet to evaluate the overall reconstruction approach. 
However, since ShapeNet features already-segmented objects, it is not suitable for evaluating silhouette guidance. For that, we use the Pix3D dataset~\cite{sun2018pix3d}, which contains real images of occluded objects with aligned 3D shape ground truth.  
We consider the following two standard metrics for 3D point cloud reconstruction:
\begin{enumerate}
\vspace{-2mm}
    \item Chamfer distance~(CD), defined in Eq.~\ref{loss:3d}. 
    \vspace{-2mm}
    \item Earth mover's distance~(EMD), defined as follows:
    \vspace{-2mm}
    \begin{align}
        d_{\mathit{EMD}}(P, \hat{P})=\min_{\phi:P\rightarrow{\hat{P}}}\frac{1}{|P|}\sum_{p\in P}\|p-\phi({p})\|
    \end{align}
where $\phi:P\rightarrow{\hat{P}}$ denotes a bijection that minimizes the average distance between corresponding points. Since $\phi$ is expensive to compute, we follow the approximation solution as in Fan~\etal~\cite{fan2017point}.
\end{enumerate}

For silhouette completion, we evaluate our approach on the synthetic DYCE dataset~\cite{ehsani2018segan} and compare with the state-of-the-art. DYCE contains segmentation mask for both visible and occluded region in photo-realistic indoor scenes. We also report performance on the Pix3D real dataset since we perform silhouette guided reconstruction on Pix3D. We use the evaluation metric of 2D IoU between the predicted and the ground truth complete silhouette. For detailed comparison, we consider the 2D IoU of the visible potion, the invisible portion and the complete silhouette.

To verify the impact of silhouette guidance, we compare with the following three baselines:
\begin{enumerate}
\vspace{-2mm}
    \item Without silhouette guidance~(ours w/o seg); 
    \vspace{-2mm}
    \item Guided by visible silhouette~(ours w/ vis seg). We use the predicted visible silhouette~(semantic segmentation by Mask-RCNN) to guide reconstruction;
    \vspace{-2mm}
    \item Guided by ground truth complete silhouette~(ours w/ gt full seg). This shows the upper bound performance of our approach.
    
\end{enumerate}

\begin{figure}[t]
\begin{center}
\includegraphics[width=1.0\linewidth]{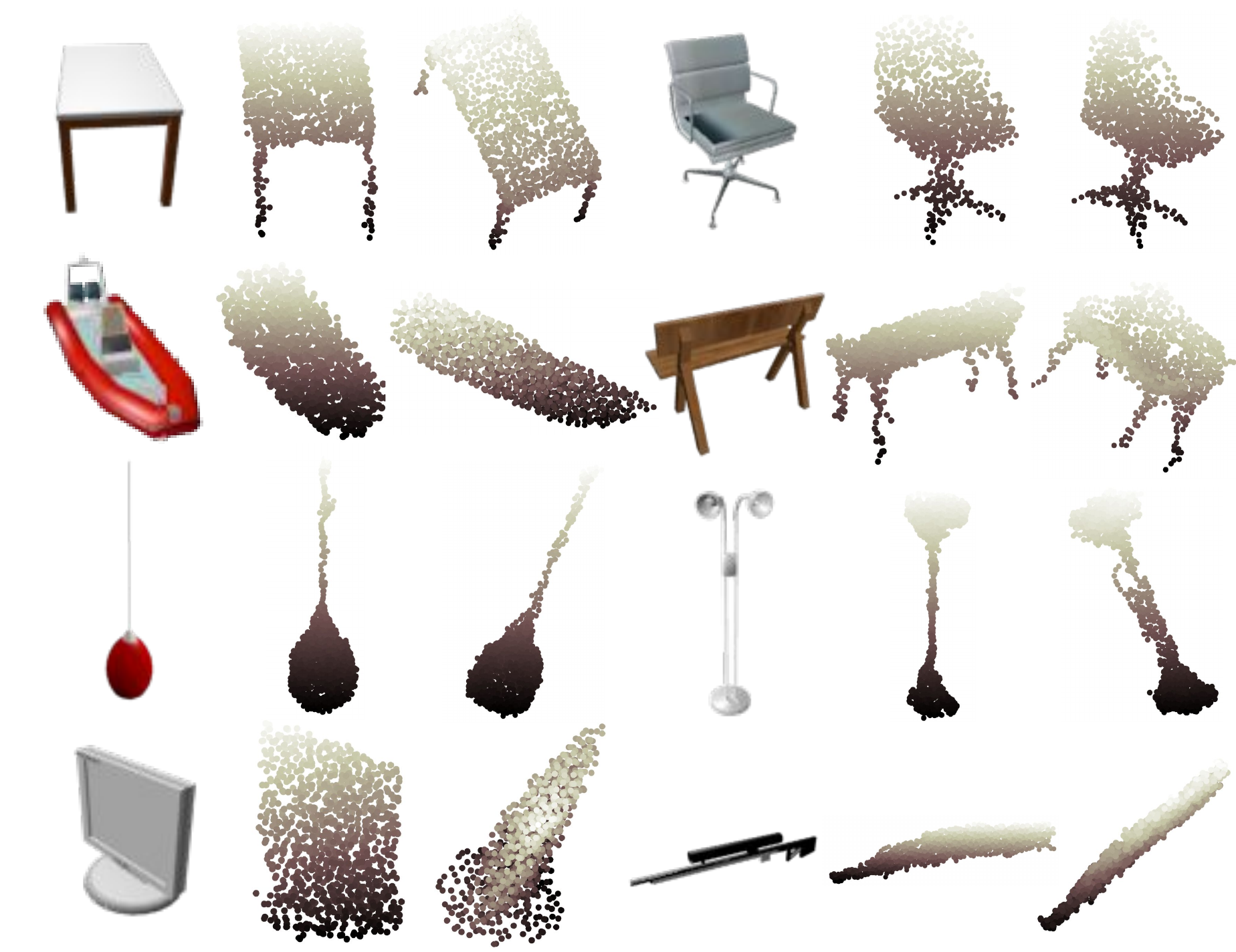}
\end{center}
\vspace{-3mm}
   \caption{Qualitative results of point cloud reconstruction on ShapeNet dataset. Each from left to right: input RGB image, reconstruction in viewer-centered coordinates and a novel view of the predicted 3D shape. Our approach is able to reconstruct thin structures and smooth surfaces. Best viewed in color.}
   \vspace{-2mm}
\label{fig:svr}
\end{figure}

\subsection{Implementation Details}
We implement our network using PyTorch, train and test on a single NVIDIA Titan X GPU with 12GB memory. A single forward pass of the network takes 15.2 ms. The point cloud refinement step is C++ based and takes around 1s on a Linux machine with Intel Xeon 3.5G Hz in CPU mode. 

To train our reconstruction approach on ShapeNet, we follow the train-test split defined by Choy~\etal~\cite{choy20163d}. The dataset consists 13 objects classes. Each object has 24 randomly selected views rendered in 2D. We randomly select 10\% of the shapes from each class of the train set to form the validation set. The viewer-centered point clouds ground truth is generated by Wang~\etal~\cite{wang2018pixel2mesh}. 
Since ShapeNet objects are non-occluded, we train our approach without silhouette guidance. 

To train the network with silhouette guidance, we construct a synthetic occlusion dataset~(Sec.~\ref{text:occ_network}) based on ShapeNet and add real background from LSUN dataset. We fine-tune the network we previously trained on ShapeNet on our generated synthetic occlusion data. We train the baseline network having predicted visible silhouette as input with the ground truth visible silhouette generated from ShapeNet. We train the baseline network having no silhouette guidance with RGB images as input only. All configurations use the same training settings. 
It's worth noting that we do not train reconstructions on Pix3D in any of our experiments to avoid the classification/retrieval problem as discussed by Tatarchenko~\etal~\cite{tatarchenko2019single}.

For silhouette completion, since Pix3D has fewer images, we first train our approach on the synthetic DYCE dataset. We use DYCE's official train-val split and use ground truth visible 2D silhouette as input. On Pix3D, we use Mask-RCNN to detect and segment the visible silhouette of the object in each image, then perform completion upon the segmentation. We obtain valid detections with correctly detected object class and a 2D IoU $>0.5$ compared to the ground truth bounding box. We use 5-fold cross validation to fine-tune the silhouette completion network pre-trained on DYCE. We further split out 10\% val data from each train split in each fold to tune network parameters. Note that images in Pix3D with the same 3D ground truth model are in the same split of either train, val or test. 



\begin{table}
\begin{center}
\resizebox{0.49\textwidth}{!}{
\begin{tabular}{cccc|ccc}
\hline
\multirow{2}{*}{Category} & \multicolumn{3}{c|}{CD} & \multicolumn{3}{|c}{EMD}\\
\cline{2-7}
 & PSG & Pixel2Mesh & Ours & PSG & Pixel2Mesh & Ours\\
\hline
plane & 0.430 & 0.477 & \textbf{0.386}& \textbf{0.396} & 0.579 & 0.527\\
bench & 0.629 & 0.624 & \textbf{0.436} & 1.113 & 0.965 &\textbf{0.815}\\
cabinet &0.439 & 0.381 & \textbf{0.373}& 2.986 & 2.563 & \textbf{2.147}\\
car & 0.333 & \textbf{0.268} & 0.308& 1.747 & \textbf{1.297} & 1.306\\
chair &0.645 & 0.610 & \textbf{0.606}& 1.946 & 1.399 & \textbf{1.257}\\
monitor &0.722 & 0.755 & \textbf{0.501} & 1.891 & 1.536 & \textbf{1.314}\\
lamp & 1.193 & 1.295 & \textbf{0.969}& 1.222 & 1.314 & \textbf{1.007}\\
speaker & 0.756 & 0.739 & \textbf{0.632}& 3.490 & 2.951 & \textbf{2.441}\\
firearm & \textbf{0.423} & 0.453 & 0.463& \textbf{0.397} & 0.667 & 0.572\\
couch & 0.549 & 0.490 & \textbf{0.439}& 2.207 & 1.642 & \textbf{1.536}\\
table & 0.517 & \textbf{0.498} & 0.589 & 2.121 & 1.480 & \textbf{1.340}\\
cellphone & 0.438 & 0.421 & \textbf{0.332}& 1.019 & 0.724 & \textbf{0.674}\\
watercraft & 0.633 & 0.670 & \textbf{0.478}& 0.945 & 0.814 & \textbf{0.730}\\
\hline
mean &0.593 & 0.591 & \textbf{0.501} & 1.653 & 1.380 & \textbf{1.205}\\
\hline
\end{tabular}}
\end{center}
\vspace{-2mm}
\caption{Viewer-centered single image shape reconstruction performance compared with the state-of-the-art on ShapeNet. We report both the Chamfer distance~(CD, left) and the Earth mover's distance~(EMD, right). 
}\label{tab:pix2mesh}
\vspace{-1.5mm}
\end{table}

\subsection{Single Image 3D Reconstruction without Occlusion}\label{exp:svr}


We show in Fig.~\ref{fig:svr} sample qualitative results of our reconstructions on ShapeNet. We report in Tab.~\ref{tab:pix2mesh} our quantitative comparison with the state-of-the-art viewer-centered reconstruction approaches. PSG~\cite{fan2017point} generates point clouds and Pixel2Mesh~\cite{wang2018pixel2mesh} produces meshes. We sample the same 2466 points as PSG and Pixel2Mesh for fair comparison. Our approach outperforms the two methods on both metrics. 

\textbf{Ablation study.} We show in Tab.~\ref{tab:abla} our performance with different configurations: without both surface-based refinement step and 2D reprojection loss, without refinement step only, without 2D reprojection loss only and our full approach. We observe the improvement with 2D reprojection loss on both CD and EMD. The improvement with 2D loss is less significant when we have the surface-based refinement step, showing the refinement step mitigates some problems with point cloud quality that are otherwise remedied by training with the reprojection loss. 
The refinement step increases Chamfer distance, mainly due to the smoothed out small and sparse point pieces that model the thin and complex shape structure like railings or handles (Fig.~\ref{fig:svr}, 1st row, 2nd column), and the enhanced error point predictions by connecting sparse point sets (Fig.~\ref{fig:svr}, 3rd row, 2nd column). 
It's also worth noting that, even without the post-refinement step, our approach achieves a better CD than PSG and a slightly worse EMD, proving the superiority of our network architecture.

\begin{table}
\begin{center}
\resizebox{0.45\textwidth}{!}{
\begin{tabular}{|c|c|c|}
\hline
Method & CD & EMD\\
\hline
\hline
Ours w/o surface refine~\& w/o 2D proj loss & 0.398& 1.784\\
Ours w/o surface refine& \textbf{0.389}& 1.660\\
Ours w/o 2D proj loss & 0.502& 1.220\\
Ours Full&0.501 & \textbf{1.205}\\
\hline
\end{tabular}}
\end{center}
\vspace{-2mm}
\caption{Ablation study. We evaluate our performance on ShapeNet based on CD~(left) and EMD~(right) with different configurations. 
Our full approach seeks for a balanced performance on both two metrics. 
}\label{tab:abla}
\vspace{-2mm}
\end{table}


\begin{figure*}
\begin{center}
\includegraphics[width=1.0\linewidth]{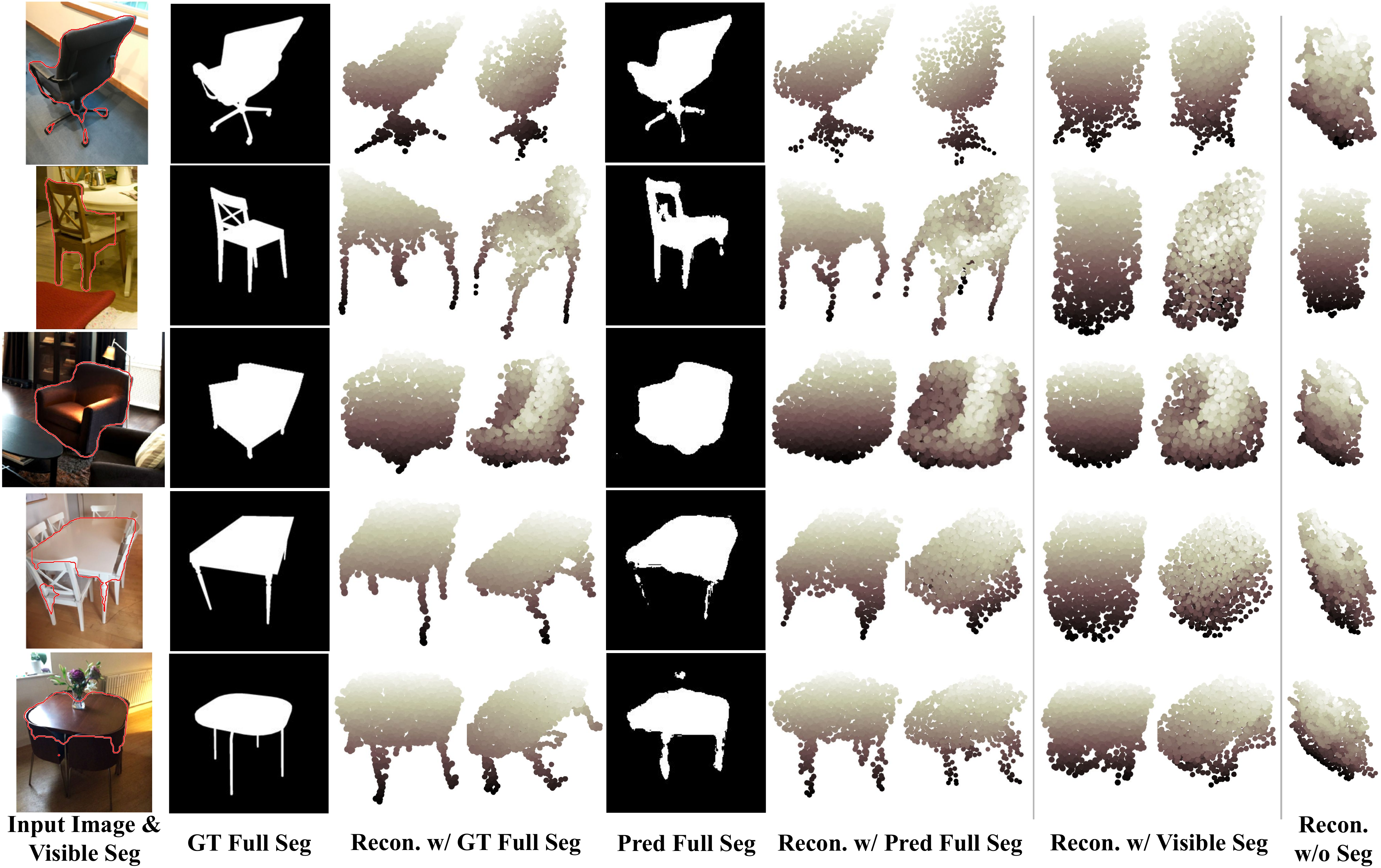}
\end{center}
\vspace{-2.5mm}
   \caption{Qualitative results on Pix3D dataset. We show in each row from left to right: input RGB image with predicted visible silhouette obtained by Mask-RCNN~(outlined in red), ground truth complete silhouette, reconstruction guided by ground truth complete silhouette in two views~(viewer-centered and a novel view), our predicted complete silhouette, reconstruction guided by predicted complete silhouette, reconstruction guided by visible silhouette and reconstruction without silhouette guidance. The first row shows a non-occluded object, and other rows show occluded objects. 
   Best viewed in color.}
\vspace{-3mm}
\label{fig:quali-occ}
\end{figure*}


\begin{table}
\begin{center}
\begin{tabular}{|c|c|c|}
\hline

Method & CD & EMD\\
\hline
3D-LMNet~\cite{mandikal20183dlmnet} &\textbf{5.40} &7.00 \\
Ours & 5.54&\textbf{5.93}\\ 
\hline
\end{tabular}
\end{center}
\vspace{-2mm}
\caption{Comparison with the state-of-the-art object-centered point cloud reconstruction approach on ShapeNet, reported in both CD~(left) and EMD~(right). Although our method is trained on a harder viewer-centered prediction, our method achieves a much better EMD and a slightly worse CD compared to 3D-LMNet.}\label{tab:lmnet}
\vspace{-2.5mm}
\end{table}

\begin{table}
\begin{center}
\begin{tabular}{|c|c|c|c|}
\hline
Method & Full & Visible & Occluded\\ 
\hline
\hline
SeGAN~\cite{ehsani2018segan} & 76.4 & 63.9 & 27.6\\
Ours~(ResNet-18) & 82.8 & 82.9 & 33.9\\
Ours~(ResNet-50) & \textbf{84.3} & \textbf{83.4} & \textbf{36.2}\\
\hline
\end{tabular}
\end{center}
\vspace{-2mm}
\caption{Object silhouette completion performance compared with the state-of-the-art SeGAN on DYCE dataset. We report the 2D IoU of visible, occluded and complete silhouette. ``Ours~(ResNet-18)'' use the same encoder as SeGAN.
}\label{tab:segan}
\vspace{-2.5mm}
\end{table}

\textbf{Comparison with object-centered approach.} Tab.~\ref{tab:lmnet} shows our comparison with the state-of-the-art object-centered point cloud reconstruction approach: 3D-LMNet~\cite{mandikal20183dlmnet} on ShapeNet. We follow the evaluation procedure as 3D-LMNet, sample 1024 points and re-scale our prediction~(and ground truth) to be zero-centered and unit length of 1, then perform ICP to fit to the ground truth. Although our approach targets the more difficult task of joint shape prediction and view point estimation, we achieve a much better EMD and only a slightly worse CD. Note that the reported CD and EMD are of different scales compared to that in Tab.~\ref{tab:pix2mesh}, this is because 3D-LMNet takes a squared value when computing CD and the ground truth is re-sized. 




\subsection{Silhouette Completion}\label{exp:sc}

Tab.~\ref{tab:segan} shows the comparison with the state-of-the-art silhouette completion approach SeGAN~\cite{ehsani2018segan} on DYCE test set. 
SeGAN uses ResNet-18 encoder, our approach with ResNet-18 outperforms SeGAN, due to our better up-sampling based decoder that predicts better region beyond occlusion, and the skip connection architecture that preserves the visible region. 
We use ResNet-50 encoder which yields the best performance to complete silhouettes for the downstream reconstruction task. 

Tab.~\ref{tab:sil-pix3d} reports our silhouette completion performance on Pix3D. Our approach achieves a better performance than the Mask-RCNN baseline~(visible region). We show better performance by fine-tuning our completion approach on Pix3D~(``syn+real'' v.s. ``syn'' for training data).  

\begin{table}
\begin{center}
\resizebox{0.49\textwidth}{!}{
\begin{tabular}{|c|c|c|c|c|c|c|c|}
\hline
\multirow{2}{*}{Method} & \multirow{2}{*}{\begin{tabular}{c}Training\\ data\end{tabular}} & \multicolumn{3}{c|}{Occluded} & \multicolumn{3}{c|}{Non-occluded}\\
\cline{3-8}
&& sofa & chair & table & sofa & chair & table\\
\hline
Mask-RCNN & real & 84.34 & 59.05 &60.14& 91.99 & 69.96 & 60.88\\
Ours&syn&87.58 & \textbf{59.61} & 58.12 & 92.02 &69.88 & \textbf{64.94}\\
Ours & syn+real &\textbf{88.56} & 59.25 & \textbf{68.83} & \textbf{92.19} & \textbf{72.01} & 56.90\\
\hline
\end{tabular}}
\end{center}
\vspace{-1.5mm}
\caption{Silhouette completion performance on Pix3D dataset. We report the 2D IoU between the predicted and the ground truth complete silhouette for occluded and non-occluded objects. 
}\label{tab:sil-pix3d}
\vspace{-2.5mm}
\end{table}

\begin{table}
\begin{center}
\resizebox{0.49\textwidth}{!}{
\begin{tabular}{|c|c|c|c|c|c|c|}
\hline
\multirow{2}{*}{Method} & \multicolumn{3}{c|}{CD} & \multicolumn{3}{c|}{EMD}\\
\cline{2-7}
& sofa & chair & table & sofa & chair & table\\
\hline
Ours w/o seg &15.54&17.96& 24.35& 16.63&16.51&22.56 \\
Ours w/ pred vis seg&9.15 &  13.20 & 17.96 &9.29 &13.38&17.81\\ 
Ours w/ pred full seg& \textbf{8.70} & \textbf{13.14} & 
\textbf{16.50}&\textbf{8.81}&\textbf{13.04}&\textbf{16.36}\\
\hline
Ours w/ gt full seg&8.27 & 10.16 & 11.36& 8.11&10.47& 11.44 \\
\hline
\end{tabular}}
\end{center}
\vspace{-1.5mm}
\caption{Quantitative results for reconstructing \textit{occluded} objects in the Pix3D dataset. We report both CD~(left) and EMD~(right).
}\label{tab:occ}
\vspace{-2.5mm}
\end{table}


\begin{table}
\begin{center}
\resizebox{0.49\textwidth}{!}{
\begin{tabular}{|c|c|c|c|c|c|c|}
\hline
\multirow{2}{*}{Method} & \multicolumn{3}{c|}{CD} & \multicolumn{3}{c|}{EMD}\\
\cline{2-7}
& sofa & chair & table & sofa & chair & table\\
\hline
Ours w/o seg& 12.62 & 16.00& 20.65& 13.18& 15.44& 19.92\\
Ours w/ pred vis seg&  8.75&11.34&15.55& 8.66 & 11.84 & 15.75\\
Ours w/ pred full seg& \textbf{8.42}&\textbf{10.82} & \textbf{13.65}& \textbf{8.40} & \textbf{11.13} & \textbf{13.85}\\
\hline
Ours w/ gt full seg& 8.24&9.21&10.50& 8.18 & 9.66 & 11.44\\
\hline
\end{tabular}}
\end{center}
\vspace{-1.5mm}
\caption{Quantitative results for reconstructing \textit{non-occluded} objects in the Pix3D dataset. We report CD~(left) and EMD~(right).
}\label{tab:notocc}
\vspace{-2mm}
\end{table}

\subsection{Robustness to Occlusion}\label{exp:occ}

In Fig.~\ref{fig:quali-occ} we show our qualitative performance on Pix3D for the three object classes that co-occur in ShapeNet. Tab.~\ref{tab:occ} and Tab.~\ref{tab:notocc} present our quantitative performance of reconstructing occluded objects and non-occluded objects respectively. For evaluation, we use the ground truth point cloud provided by Mandikal~\etal~\cite{mandikal20183dlmnet} and sample the same number of points from our approach for evaluation. Since Pix3D evaluates object-centered reconstruction, we rotate each ground truth shape to have the viewer-centered orientation w.r.t. camera. We then re-scale both ground truth and our prediction to be zero-centered and unit-length, and perform ICP with translation only. 
We follow the officially provided evaluation metrics of Chamfer distance and Earth mover's distance by Sun~\etal~\cite{sun2018pix3d}. We compare the performance of our silhouette guided reconstruction ``Ours w/~pred full seg'' with three baselines: without silhouette guidance, guided with predicted visible silhouette from Mask-RCNN, and guided with ground truth complete silhouette. Compared to the qualitative results on ShapeNet~(Fig.~\ref{fig:svr}), we see the challenges of 3D reconstruction given real images due to occlusion and complex background. Using ground truth complete silhouette can make the network be robust to occlusion. Without silhouette guidance, the prediction is difficult because the network does not know whether or where an object is occluded and what to reconstruct; and the network faces the challenges of synthetic to real, since our reconstruction network is only trained on synthetic dataset. Our proposed silhouette guidance is able to bridges the gap between synthetic and real, referring to the large performance boosts from ``Ours w/o seg'' to other rows. With the guidance of predicted complete silhouettes, our approach is able to narrow down the performance gap between occluded and non-occluded objects, and outperforms the approach with predicted visible silhouettes. 
\section{Conclusion}
We propose a method to reconstruct the complete 3D shape of an object from a single RGB image, with robustness to occlusion. Our point cloud reconstruction approach achieves the state-of-the-art with the major improvement by the surface-based refinement step. 
We show that, when provided with input ground truth silhouettes, the shape prediction performance is nearly as good for occluded as for non-occluded objects.  Using the predicted silhouette also yields large improvements for both occluded and non-occluded objects, indicating that providing an explicit foreground/background separation for the object in RGB images is helpful. 



\section*{Acknowledgements}
\vspace{-1.5mm}
This research is supported in part by NSF award 14-21521 and ONR MURI grant N00014-16-1-2007.

{\small
\bibliographystyle{ieee}
\bibliography{egbib}
}

\appendix 
\section{Illustration of 2D Reprojection Loss}\label{appx:proj}
\begin{figure}
\begin{center}
\includegraphics[width=0.6\linewidth]{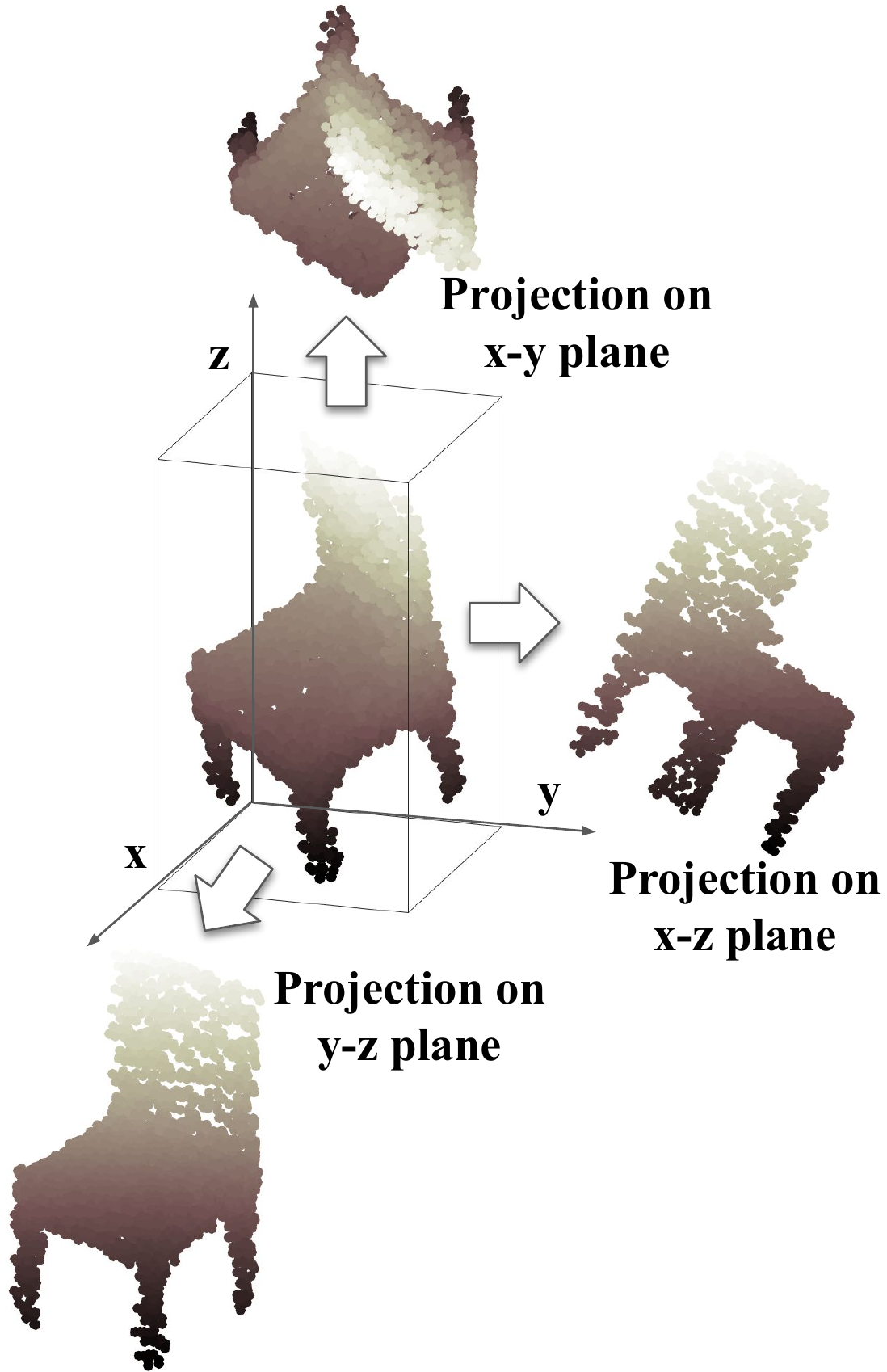}
\end{center}
\vspace{-2.5mm}
   \caption{Illustration of our multi-view 2D reprojection loss.}
\vspace{-3mm}
\label{fig:2d-proj}
\end{figure}

We show in Fig.~\ref{fig:2d-proj} a detailed description of our 2D reprojection loss from multiple views. For each predicted object, we consider three orthogonal projections based on right-handed coordinates: projection on the y-z plane, projection on the x-z plane and projection on the x-y plane. Note that the projection on the y-z plane is equal to the projection on the image plane~(frontal view). 

\section{Reconstruction Performance with Silhouette Guidance on ShapeNet}
Tab.~\ref{tab:abla2} shows our reconstruction performance with silhouette guidance on ShapeNet. Since ShapeNet images~\cite{choy20163d} use white background and have no foreground occlusion, we obtain silhouettes with image intensity $< 1$~(given image value ranges between $[0,1]$) for both training and testing. With silhouette guidance the overall performance drops slightly. This is because the input image with a white background provides sufficient evidence for the network to capture 3D shape feature, and due to the ambiguity of similar silhouettes from completely different shapes, additional silhouette guidance will confuse the network.

\begin{table}
\begin{center}
\resizebox{0.49\textwidth}{!}{
\begin{tabular}{ccc|cc}
\hline
\multirow{2}{*}{Category} & \multicolumn{2}{c|}{CD} & \multicolumn{2}{|c}{EMD}\\
\cline{2-5}
 & Ours & Ours w/~silhouette & Ours & Ours w/~silhouette\\
\hline
plane & \textbf{0.386}& 0.395& \textbf{0.527} & 0.552\\
bench & \textbf{0.436} & 0.450&\textbf{0.815} & 0.826\\
cabinet & 0.373&\textbf{0.365} & \textbf{2.147}& 2.154\\
car & 0.308& \textbf{0.301}& 1.306 &\textbf{1.302}\\
chair & \textbf{0.606}&  0.637& \textbf{1.257}&1.275\\
monitor & \textbf{0.501} & 0.519& \textbf{1.314}&1.338\\
lamp & \textbf{0.969}& 0.991& 1.007&\textbf{1.005}\\
speaker & 0.632&\textbf{0.626}  & 2.441&\textbf{2.408}\\
firearm & \textbf{0.463}& 0.480 & \textbf{0.572} &0.580\\
couch & \textbf{0.439}& 0.455 & \textbf{1.536} &1.568\\
table & \textbf{0.589} & 0.604 & 1.340 &\textbf{1.338}\\
cellphone & \textbf{0.332}& 0.339 & \textbf{0.674}&0.687\\
watercraft & \textbf{0.478}& 0.492& \textbf{0.730} &0.747\\
\hline
mean &\textbf{0.501} & 0.512 & \textbf{1.205} & 1.214\\
\hline
\end{tabular}}
\end{center}
\vspace{-2mm}
\caption{Viewer-centered single image shape reconstruction performance on ShapeNet. We report both Chamfer distance~(CD, left) and Earth mover's distance~(EMD, right). We compare our approach to ours using additional silhouette guidance. For datasets like ShapeNet that have no foreground occlusion and a white background, silhouette guidance can hardly help.}\label{tab:abla2}
\vspace{-1.5mm}
\end{table}



\section{Detailed Quantitative Analysis on Pix3D}
We evaluate the performance of reconstructing occluded objects under different occlusion ratio. Pix3D~\cite{sun2018pix3d} has detailed ground truth label of ``slightly occluded'' and ``highly occluded''. We report reconstruction performance classified by these two labels as follows. 

\begin{table}
\begin{center}
\resizebox{0.49\textwidth}{!}{
\begin{tabular}{|c|c|c|c|c|c|c|c|}
\hline
\multirow{2}{*}{Method} & \multirow{2}{*}{\begin{tabular}{c}Training\\ data\end{tabular}} & \multicolumn{3}{c|}{Slightly Occluded} & \multicolumn{3}{c|}{Highly Occluded}\\
\cline{3-8}
&& sofa & chair & table & sofa & chair & table\\
\hline
Mask-RCNN & real & 89.10 & 63.57 & 61.02 & 78.79 & 56.87 & 59.56\\
Ours&syn& 90.18 & 63.88 & 60.09 & 84.54 & \textbf{57.55} & 56.83\\
Ours & syn+real & \textbf{90.45} & \textbf{63.89} & \textbf{66.45} & \textbf{86.36} & 57.01 & \textbf{60.47}\\
\hline
\end{tabular}}
\end{center}
\vspace{-1.5mm}
\caption{Silhouette completion performance of occluded object in the Pix3D dataset. We report 2D IoU between the predicted and the ground truth complete silhouette. We split objects into categories of ``slightly occluded'' and ``highly occluded'' defined by Pix3D. 
}\label{tab:sil-pix3d-d}
\vspace{-2.5mm}
\end{table}

\textbf{Silhouette completion.} Tab.~\ref{tab:sil-pix3d-d} reports the performance of completing slightly occluded and highly occluded silhouettes respectively. 
Completion is easier for slightly occluded objects than highly occluded objects. Our silhouette completion network fine-tuned on Pix3D shows better performance than Mask-RCNN baseline. For sofas, our approach has similar performance for slightly occluded and highly occluded objects. For chairs, completion is difficult due to the complex shape variations beyond occlusions, but we still slightly outperform Mask-RCNN. For tables, our approach is able to have much better completion compared to Mask-RCNN baseline for slightly occluded objects than highly occluded objects. This is because the visible segmentation obtained by Mask-RCNN is not accurate, making silhouette completion difficult.

\textbf{Silhouette guided point clouds reconstruction.} Tab.~\ref{tab:socc-d} and Tab.~\ref{tab:occ-d} show quantitative results for reconstructing slightly occluded and highly occluded objects in the Pix3D dataset respectively. For all methods, reconstructing slightly occluded chairs and tables is easier than reconstructing highly occluded chairs and tables. For sofas, reconstruction performance is less affected by the occlusion ratio. Overall, our method with predicted complete silhouette performs better than our method guided by visible silhouettes, except for highly occluded chairs that are difficult to predict due to their complex shape beyond occlusion. 

\begin{table}
\begin{center}
\resizebox{0.49\textwidth}{!}{
\begin{tabular}{|c|c|c|c|c|c|c|}
\hline
\multirow{2}{*}{Method} & \multicolumn{3}{c|}{CD} & \multicolumn{3}{c|}{EMD}\\
\cline{2-7}
& sofa & chair & table & sofa & chair & table\\
\hline
Ours w/o seg & 15.34& 17.82& 22.84& 16.40& 16.48& 21.25\\
Ours w/ pred vis seg& 9.04& 12.46 & 16.28 & 8.99& 12.75& 16.54\\ 
Ours w/ pred full seg& \textbf{8.58} & \textbf{11.83} & \textbf{13.92}
& \textbf{8.49}& \textbf{12.00}& \textbf{13.96}\\
\hline
Ours w/ gt full seg& 8.26& 9.58 & 8.77& 7.97& 9.98& 9.59 \\
\hline
\end{tabular}}
\end{center}
\vspace{-1.5mm}
\caption{Quantitative results for reconstructing \textit{slightly occluded} objects in the Pix3D dataset. We report both CD~(left) and EMD~(right).
}\label{tab:socc-d}
\vspace{-2.5mm}
\end{table}

\begin{table}
\begin{center}
\resizebox{0.49\textwidth}{!}{
\begin{tabular}{|c|c|c|c|c|c|c|}
\hline
\multirow{2}{*}{Method} & \multicolumn{3}{c|}{CD} & \multicolumn{3}{c|}{EMD}\\
\cline{2-7}
& sofa & chair & table & sofa & chair & table\\
\hline
Ours w/o seg & 15.77& 18.03& 25.33& 16.90& 16.52& 23.41\\
Ours w/ pred vis seg& 9.27& \textbf{13.55} & 19.06 & 9.65& 13.68& 18.64\\ 
Ours w/ pred full seg& \textbf{8.85} & 13.76 &\textbf{18.19} 
&\textbf{9.19} &\textbf{13.53} &\textbf{17.93} \\
\hline
Ours w/ gt full seg& 8.28& 10.44 &13.05 & 8.27&10.71 &12.65  \\
\hline
\end{tabular}}
\end{center}
\vspace{-1.5mm}
\caption{Quantitative results for reconstructing \textit{highly occluded} objects in the Pix3D dataset. We report both CD~(left) and EMD~(right).
}\label{tab:occ-d}
\vspace{-2.5mm}
\end{table}

\section{More Qualitative Results on Pix3D}
\begin{figure*}
\begin{center}
\includegraphics[width=1.0\linewidth]{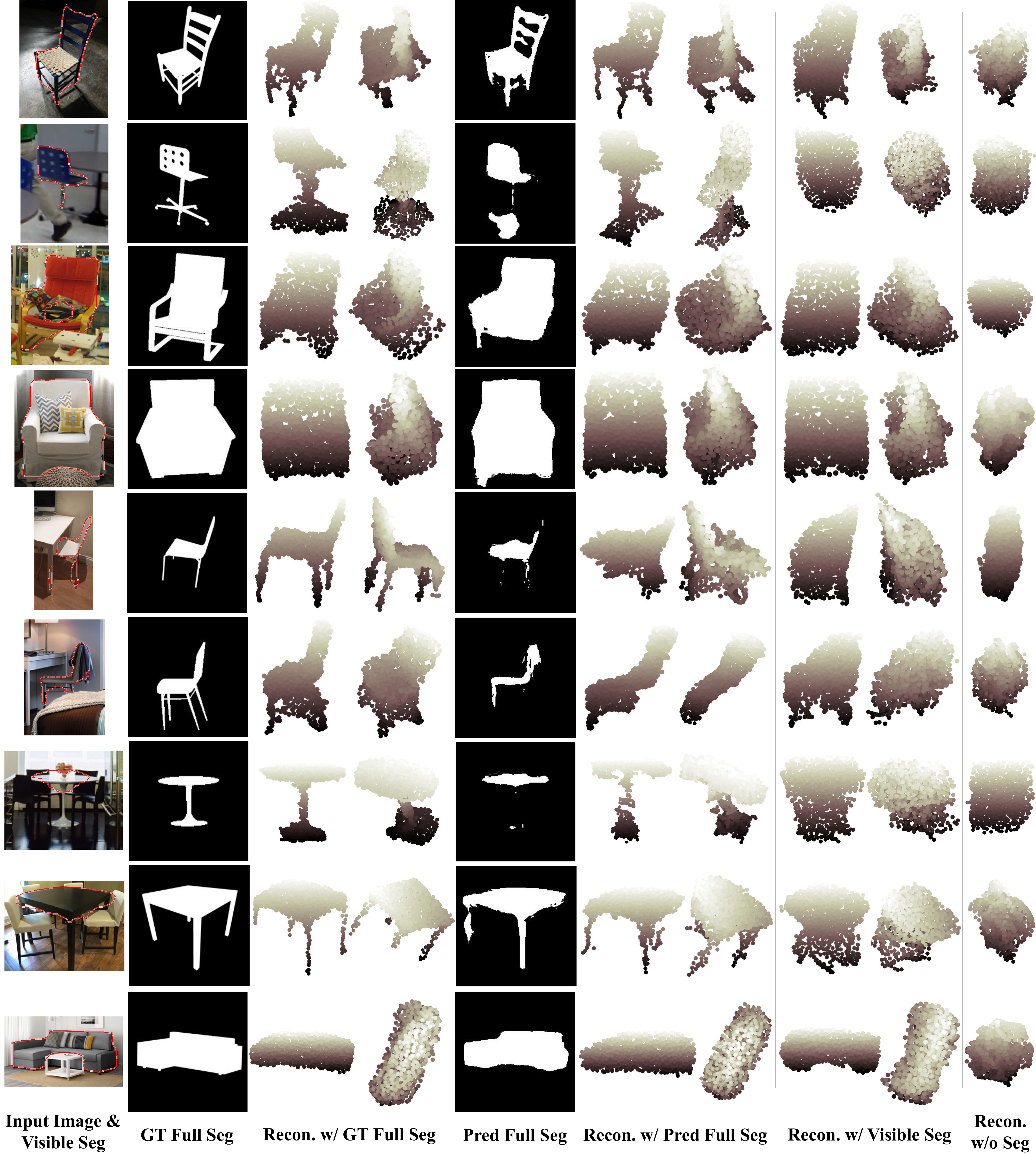}
\end{center}
\vspace{-2.5mm}
   \caption{More qualitative results on Pix3D dataset. We show in each row from left to right: input RGB image with predicted visible silhouette obtained by Mask-RCNN~(outlined in red), ground truth complete silhouette, reconstruction guided by ground truth complete silhouette in two views~(viewer-centered and a novel view), our predicted complete silhouette, reconstruction guided by predicted complete silhouette, reconstruction guided by visible silhouette and reconstruction without silhouette guidance. The first row shows a non-occluded object, and other rows show occluded objects. Best viewed in color.}
\vspace{-3mm}
\label{fig:quali-occ-2}
\end{figure*}

\begin{figure*}
\begin{center}
\includegraphics[width=1.0\linewidth]{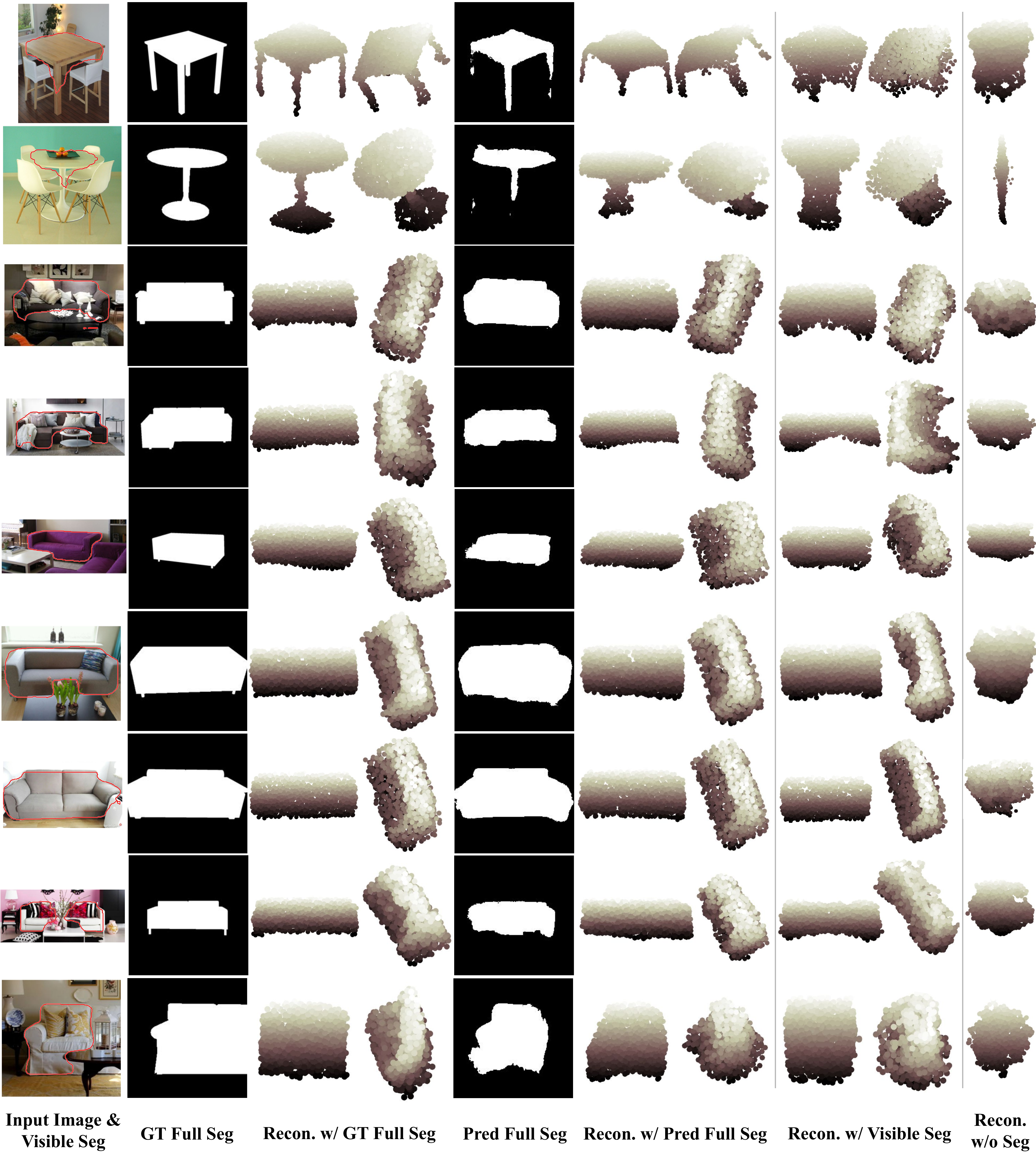}
\end{center}
\vspace{-2.5mm}
   \caption{More qualitative results on Pix3D dataset. We show in each row from left to right: input RGB image with predicted visible silhouette obtained by Mask-RCNN~(outlined in red), ground truth complete silhouette, reconstruction guided by ground truth complete silhouette in two views~(viewer-centered and a novel view), our predicted complete silhouette, reconstruction guided by predicted complete silhouette, reconstruction guided by visible silhouette and reconstruction without silhouette guidance. Best viewed in color.}
\vspace{-3mm}
\label{fig:quali-occ-1}
\end{figure*}

We show in Fig.~\ref{fig:quali-occ-2} and Fig.~\ref{fig:quali-occ-1} more qualitative results on Pix3D dataset.


\end{document}